%% file: main.tex
\documentclass[sigconf]{acmart}

\usepackage{booktabs} 
\usepackage{fixltx2e}
\usepackage{hyperref}
\usepackage[utf8]{inputenc} 
\usepackage[T1]{fontenc}
\usepackage{amsmath}
\usepackage{amsfonts}
\usepackage{dsfont}
\usepackage{subfigure} 

\setcopyright{rightsretained}

\acmDOI{10.475/123_4}

\acmISBN{123-4567-24-567/08/06}

\acmConference[SIGIR 2019]{ACM Woodstock conference}{July 1997}{El
  Paso, Texas USA}
\acmYear{1997}
\copyrightyear{2016}

\acmArticle{4}
\acmPrice{15.00}

\begin{document}
\title{A Compact Deep Learning Model for Face Spoofing Detection}
\subtitle{Wide and Deep Features for Face Presentation Attack Detection}

\author{Seyedkooshan Hashemifard, Mohammad Akbari}
\affiliation{%
  \institution{Amirkabir University of Technology}
}







\renewcommand{\shortauthors}{Anonymous.}

\newtheorem{observation} {\textbf{Observation}}
\newcommand{\paratitle}[1]{\vspace{1.5ex}\noindent \textbf{#1}}

\newcommand{\mohammad}[1]{{\bf \color{red} [[Mohammad says: #1']]}}
\newcommand{\patrycja}[1]{{\bf \color{blue} [[Patrycja says ``#1'']]}}

\begin{abstract}
In recent years, face biometric security systems are rapidly increasing, therefore, the presentation attack detection (PAD) has received significant attention from research communities and has become a major field of research. Researchers have tackled the problem with various methods, from exploiting conventional texture feature extraction such as LBP, BSIF, and LPQ to using deep neural networks with different architectures. Despite the results each of these techniques has achieved for a certain attack scenario or dataset, most of them still failed to generalized the problem for unseen conditions, as the efficiency of each is limited to certain type of presentation attacks and instruments (PAI). In this paper, instead of completely extracting hand-crafted texture features or relying only on deep neural networks, we address the problem via fusing both wide and deep features in a unified neural architecture. The main idea is to take advantage of the strength of both methods to derive well-generalized solution for the problem. We also evaluated the effectiveness of our method by comparing the results with each of the mentioned techniques separately. The procedure is done on different spoofing datasets such as ROSE-Youtu, SiW and NUAA Imposter datasets.

In particular, we simultanously learn a low dimensional latent space empowered with data-driven features learnt via Convolutional Neural Network designes for spoofing detection task (i.e., deep channel) as well as leverages spoofing detection feature already popular for spoofing in frequency and temporal dimensions ( i.e., via wide channel). 

\end{abstract}

%
%



\maketitle

\input{body.tex}

\bibliographystyle{ACM-Reference-Format}
\bibliography{main.bib}

\end{document}

%% file: body.tex
\section{Introduction}
Thanks to the advancement of deep learning, face recognition has been remarkably incorporated in most biometric systems. Thus facial biometric systems are widely used in various applications, including mobile phone authentication, access control and face payment~\cite{kunda2018survey,galterio2018review}. Face-spoofing attacks,in which a spoofed face is presented to the biometric system in an attempt to be authenticated, are becoming a inevitable threat~\cite{el2020deep,rattani2018survey}. 
Therefore, face-spoofing detection has become a critical requirements for any face recognition system to filter out fake faces~\cite{ramachandra2017presentation}. While face anti-spoofing techniques have received much attention to aim at identifying whether the captured face is genuine or fake, most face-spoofing detection techniques are biased towards a specific presentation attack type or presentation device; failing to robustly detects various spoofing scenarios. To mitigate this problem, we aim at developing a generalizable face-spoofing framework which able to accurately identify various spoofing attacks and devices.



Face anti-spoofing techniques have received much attention and several anti-spoofing approaches have been introduced in retrospective studies~\cite{ramachandra2017presentation,galbally2014biometric}.
Traditional image-based approaches focus on image quality and characteristics and thus employ hand-craft features, such as LBP, SIFT, HOG, and SURF, with shallow classifiers to discriminate the live and fake faces~\cite{boulkenafet2016face,de2012lbp,maatta2011face}. These hand-crafted features are limited to specific spoofing patterns, scene conditions and spoofing devices, which limits their generalization~\cite{li2016generalized}.
Lately, deep methods based on Convolutional Neural Networks (CNNs) provide an alternative way to further push the effectiveness of anti-spoofing techniques via learning a discriminate representation in an end-to-end manner~\cite{li2016original,yang2014learn}. While data-driven feature learning boost the performance of spoofing detection, these methods fail to exploit the nature of spoofing patterns, which consist of skin details, color distortion, moire patterns, glass reflection, shape deformation, etc. as they mostly build models for the current dataset and fail to generalize in cross-dataset settings. Further, they also suffer from sensitivity to lighting and illumination distortion~\cite{boulkenafet2016face} as they are built upon controlled and biased datasets. As a result, these models suffer from overfitting and poor generalizability to new patterns and environments.



While numerous machine learning models have been developed to discover artifacts in spoof images, the performance of spoofing models in practical settings is still far from perfect due to the following challenges. First, the available spoofing attack datasets are limited and bias to several environmental and capture settings as compared to other computer vision tasks such as image classification for which there exist large-scale labelled datasets, like ImageNet~\cite{deng2009imagenet}. More specifically, they collected for a specific attack scenario, e.g. reply-attack, or they collected with controlled lighting and illuminance settings with limited number of subjects, i.e., faces. Second, there exist various attack types and new attack scenarios are detected once a while, such as adversarial examples~\cite{guera2018deepfake}. Most proposed models work optimal for a specific scenario or dataset and their performance on unseen attack types (data) are unreliable. Third, current deep models are developed for semantic-rich computer vision task, e.g., object detection and image captioning, as opposed to anti-spoofing which relies on low-level features. Thus, these models fail to capture good spoofing patterns as they attempt to learn high-level semantic features. Thus, developing a task-specific model equipped with low-level discriminator is highly desired.

To tackle these challenges, we propose a dual channel neural model that directly learns optimal features to discriminate fake and genuine faces. To do so, the proposed model learns a low-dimensional latent space for face spoofing task using deep and wide channels. The former, i.e, the deep channel, learns data-driven features discriminating genuine and spoofed faces by leveraging a CNN architecture specifically designed for spoofing detection task. The later, i.e., wide channel, leverages hand-crafted feature, already popular for spoofing detection task ( in frequency, texture, and temporal dimensions) and seamlessly integrated them into the low-dimensional latent space learnt by the deep channel. The proposed framework has been extensively examined on several spoofing detection datasets to evaluate its effectiveness.


The main contributions of this paper is as follows:
\begin{itemize}
    \item Develop a well-generalized model robust to environmental changes and datasets. 
    \item Take advantage of both CNN and hand-crafted features strengths to confront with newly generated or unseen attacks. 
    \item Compare the efficiency of each methods on some of the available dataset.
    \item Propose a deep architecture fitting on low-level characteristic patterns in spoofing domain.
\end{itemize}

\section{Related Works}

The main approaches in face anti-spoofing can be broadly categorized into three groups: traditional texture discriminators, Deep Learning feature learning and motion based methods.

Texture analysis approaches are mostly effective against photo and print attacks. For instance, in~\cite{li2004live}, Li et al. employed Fourier spectra analysis and assumed fake faces are usually smaller or have fewer high frequency components in photographs compared to real faces. However, the author ignored the illumination variations. Peixoto et al.~\cite{peixoto2011face} used difference-of-Gaussian (DoG) which previously was proposed by Tan et al. in~\cite{tan2010face}, and combined it with sparse logistic regression to encounter with print attacks. Using LBP features for spoofing detection proposed by Määttä et al. in~\cite{maatta2011face} which achieved outperforming results on NUAA Imposter dataset. Razvan D. A~\cite{albu2015face}, also experiment on NUAA with random transform method. Boulkenafet et al.~\cite{boulkenafet2016face} investigated the effect of various color spaces and combination of different color-texture features such as LBP, BSIF, LBQ, SID and CoALBP. Pereira et al in~\cite{de2012lbp}, proposed a spatio-temporal texture feaure called Local Binary Patterns from Three Orthogonal Planes (LBP-TOP) which outperformed LBP based methods on Replay-Attack dataset. However, the method performance falls drastically in other datasets or more realistic cases.

Thanks to gathering of large datasets in recent years, CNN based network are able to extract discriminative features to detect spoofing attacks. For example, Yang et al.~\cite{yang2014learn} leveraged CNN to extract features from detected face in image. To consider information in other parts of image, they further feed different scales of input image to the network ($5$ different scales), from closely cropped face to further distance. Atoum et al.~\cite{atoum2017face} extracts local texture features in small patches from the faces along with estimated depth feature map for detection of the spoofing attack. LSTM-CNN architecture was leveraged to take advantage of consecutive frames in a video which was proved to be effective in the case of video replay attacks~\cite{xu2015learning}. Chen et al.~\cite{chen2019attention}, proposed a two stream convolutional neural network (TSCNN) which works on two complementary space: RGB and multi-scale retinex (MSR). space to take advantage of both texture details in RGB space and illumination invariance of MSR. Gan et al.~\cite{gan20173d}, experimented 3D CNN on public video attack datasets such as CASIA~\cite{zhang2012face} and Replay-Attack~\cite{chingovska2012effectiveness}.

Motion based methods aim to use the face dynamics, reactions and activities such as head shaking, lip movement, blinking to distinguish genuine face from fake one. Kollreider et al.~\cite{kollreider2007real} used
facial parts movement as liveness features. Pan et al.~\cite{pan2007eyeblink} used eye blink to detect spoofing attacks. In~\cite{bao2009liveness}, optical flow vectors of the face region are extracted from the video and compared between different regions. Most of the methods are designed to detect printed photo attacks and not much effective to counter video attacks. However, Tirunagari et al.~\cite{tirunagari2015detection}, applied dynamic mode decomposition (DMD) which is able to represent temporal information of spoof video in a single image.

\section{Proposed Framework}
\label{sec::framework}

\begin{figure*}
  \includegraphics[width=\linewidth]{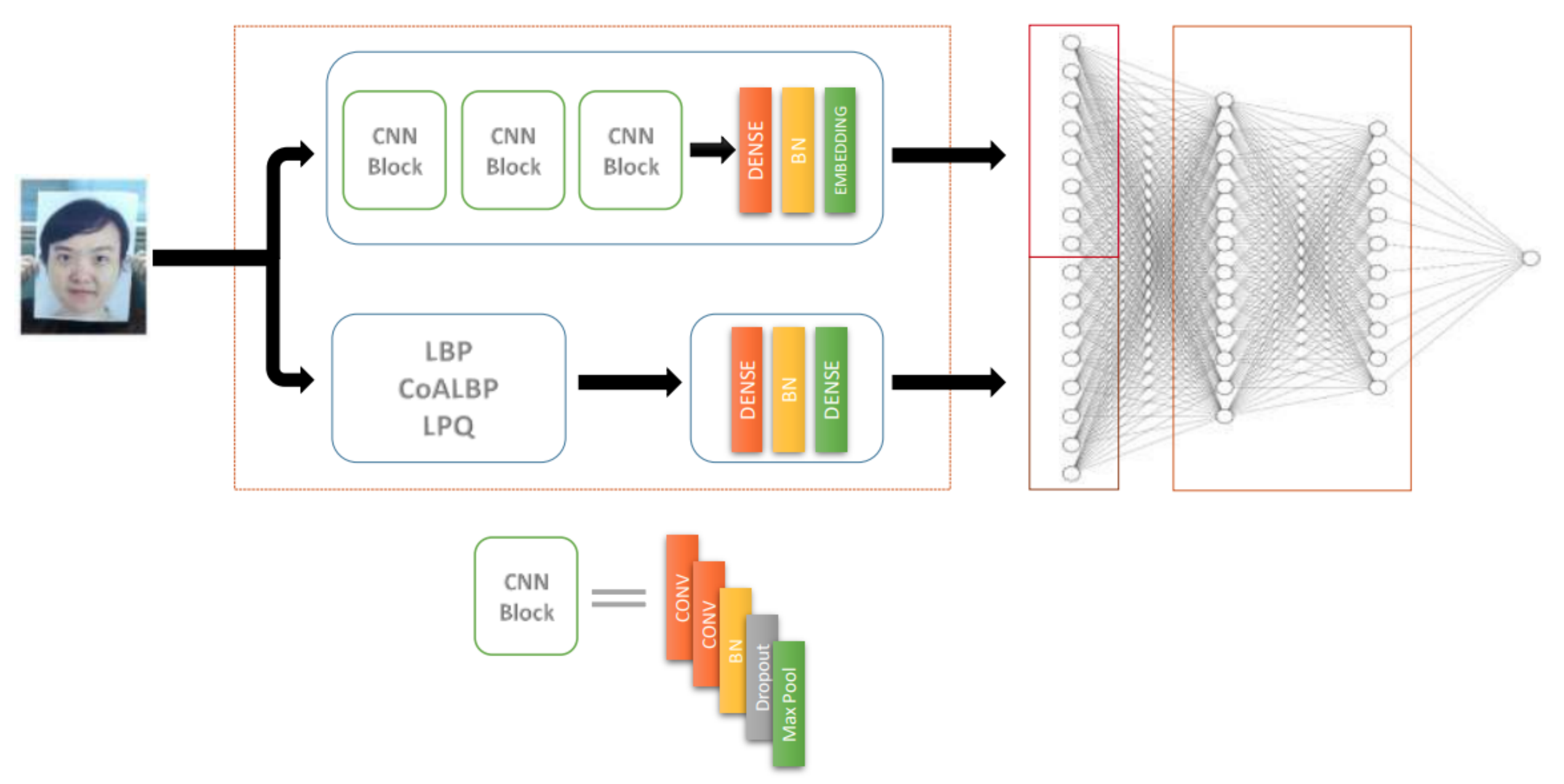}
  \caption{The schematic view of Face-Spoofing Detection Framework}
  \label{fig::framework}
\end{figure*}

We proposed a dual channel neural architecture to exploit both deep and wide features for face spoofing detection, as shown in Fig.~\ref{fig::framework}. The deep channel leverages a CNN architecture to extract discriminative patterns of spoofing from data. The designed architecture focuses to extract subtle distortions of images that represent a presentation attack. The wide channel, however, employs hand-crafted features with a shallow network to empower the model with domain-specific features known by experts. We next aggregate the extracted features of each channel into a low-dimensional latent space for final classification.

\subsection{Deep feature Learning Channel} 

In our experiments, we discovered that very deep neural models are ineffective in learning discriminative features for face spoofing detection task while employing shallower and simpler architectures can lead to better results and higher generalization.  
This can be justified by considering the nature of the problem. Face spoofing and presentation attacks usually causes illuminance distortions, reflection and soft artifacts in the image. These artifact can be considered as low level features. Thus using very deep neural model can distract the model from paying attention on these subtle alterations to some higher level semantic in images, such as face characteristics of the subjects, which explained in detail in~\cite{zeiler2014visualizing}. This suggests that neural architectures for face spoofing needs shallower CNN architectures as compared of other computer vision tasks.

This Channel contains multiple layers of convolutional and pooling followed by fully connected layer. Batch normalization and dropout are also added for regularization in order to avoid overfitting. The input consists of face image frames which are detected and cropped with margin from the dataset videos, already aligned and resized to $160 \times 160 \times 3$ in the preprocessing step. The output of the channels gives the $512$ embedding vector derived from the input face image. The deep channel network architecture is shown in Table~\ref{tbl::deep-channel-architecture}.
The embedding vector from output of the CNN channel will be passed into feature interaction learning block.


\begin{table*}
\centering
\caption{Deep channel network structure. The input and output sizes are described in rows x cols x \#filters. The layer details are specified as kernel size, stride, padding}
\label{tbl::deep-channel-architecture}
\begin{tabular}{|c|c|c|c|c|} 
\hline \hline
Layer       & Size-in    & Size-out   & Layer Details       & Parameters  \\ 
\hline
Conv1       & 160×160×3  & 158×158×32 & (3,3), S=1, P=valid & 896         \\ 
\hline
Conv2       & 158×158×32 & 156×156×32 & (3,3), S=1, P=valid & 9248        \\ 
\hline
Batch norm1 & 156×156×32 & 156×156×32 &                     & 128         \\ 
\hline
Dropout1    & 156×156×32 & 156×156×32 & Rate=0.1            & 0           \\ 
\hline
Max pool1   & 156×156×32 & 78×78×32   & (2,2), S=2, P=valid & 0           \\ 
\hline
Conv3       & 78×78×32   & 76×76×64   & (3,3), S=1, P=valid & 18,496      \\ 
\hline
Conv4       & 76×76×64   & 74×74×64   & (3,3), S=1, P=valid & 36,928      \\ 
\hline
Batch norm2 & 74×74×64   & 74×74×64   &                     & 256         \\ 
\hline
Dropout2    & 74×74×64   & 74×74×64   & Rate=0.1            & 0           \\ 
\hline
Max pool2   & 74×74×64   & 37×37×64   & (2,2), S=2, P=valid & 0           \\ 
\hline
Conv5       & 37×37×64   & 33×33×128  & (5,5), S=1, P=valid & 204,928     \\ 
\hline
Conv6       & 33×33×128  & 29×29×128  & (5,5), S=1, P=valid & 409,728     \\ 
\hline
Batch norm3 & 29×29×128  & 29×29×128  &                     & 512         \\ 
\hline
Dropout3    & 29×29×128  & 29×29×128  & Rate=0.1            & 0           \\ 
\hline
Max pool3   & 29×29×128  & 14×14×128  & (2,2), S=2, P=valid & 0           \\ 
\hline
Dense1      & 25088×1    & 512×1      & 512 neurons (relu)  & 12,845,568  \\ 
\hline
Batch norm4 & 512×1      & 512×1      &                     & 2048        \\ 
\hline
Embedding   & 512×1      & 512×1      & 512 neurons (relu)  & 262,656     \\ 
\hline
CNN Total   & \multicolumn{3}{c|}{}                         & 13,791,392  \\
\hline
\end{tabular}
\end{table*}

\subsection{Shallow Feature Exploitation} 

Retrospective studies showed color and texture features can effectively extract face spoofing patterns in videos~\cite{chingovska2012effectiveness, de2012lbp, maatta2011face}. Thus, in the wide channel, first the aligned face image is passed to a descriptor computing unit. This unit extracts color texture features, such as LBP, CoALBP and LBQ, from gray-scale, HSV and YCbCr color spaces, as showin in Table~\ref{tbl::shallow-features}. The resulting feature vectors of each descriptor shows various aspect of distortions in videos. These features are then concatenated with each other to form the total feature vector. The vector will be passed to the shallow channel to learn the embedding of the videos. The wide channel includes two fully connected layers, each contains $512$ hidden units. The last layer is then used as the embedding vector of the descriptors and will be fed to the feature interaction block to learn a low-dimensional latent space from both channels.

\begin{table*}
\centering
\caption{The parameters of the descriptors and dimensions of the concatenated RGB and YCbCr feature vectors used in our experiments}
\label{tbl::shallow-features}
\begin{tabular}{|c|c|c|} 
\hline
\textcolor[rgb]{0.133,0.133,0.133}{Descriptor} & \textcolor[rgb]{0.133,0.133,0.133}{Parameters} & \textcolor[rgb]{0.133,0.133,0.133}{Dimensions}  \\ 
\hline \hline
\textcolor[rgb]{0.133,0.133,0.133}{LBP}        & Radius R=1, Neighbors P=8                      & \textcolor[rgb]{0.133,0.133,0.133}{354}         \\ 
\hline
\textcolor[rgb]{0.133,0.133,0.133}{CoALBP}     & R=1, LBP descriptor= LBP+, B=2                 & \textcolor[rgb]{0.133,0.133,0.133}{6144}        \\ 
\hline
\textcolor[rgb]{0.133,0.133,0.133}{LPQ}        & Widows size M=3, $\alpha$=1/7                         & \textcolor[rgb]{0.133,0.133,0.133}{255}         \\
\hline
\end{tabular}
\end{table*}

\subsection{Feature Interaction Learning} 
Here, we integrate both embeddings learnt from deep and wide channels into a unified latent space and leverage it to classify the input image into genuine and spoofed face. In total, our model includes two fully-connected layers right before the output layer (described next) to allow for modelling interactions between the components of the intermediate representation, i.e., embeddings from two channels. In particular, we concatenate the outputs of the both components to form a dense vector and feed it to fully-connected layers to get more high-level and abstract features. 
Let $e_d$ and $e_w$ denotes the embedding learnt by deep and wide channels, respectively, the fully-connected layer computes,
\begin{align}
\mathbf{z} = \Phi \left( \mathbf{W} \begin{bmatrix} \mathbf{e}_d \\ \mathbf{e}_w  \end{bmatrix} + \mathbf{b} \right), 
\end{align}
where $\mathbf{W}$ and $\mathbf{b}$ are the weight vectors and bias term and $\Phi$ is the activation function for the fully connected layers. Activation function here is the ReLU non-linearity function. Here, we used two consecutive block of dense layer with $512$ hidden units for the feature interaction layer. The hypothesis behind these blocks is to learn non-linear interaction of the parameters according to the input constructed from both deep and wide embeddings, where impacts of each feature is learnt in training process. 

The problem is dealt with as a binary classification task so that the network would be either spoof or bona fide label. Therefore binary cross-entropy is utilized as the loss function of the network output layer. Table~\ref{tbl::finalblock-parameters} illustrates the final block network architecture and total parameters number.

\begin{table*}
\centering
\caption{Final block structure and total parameters of the proposed network}
\label{tbl::finalblock-parameters}
\begin{tabular}{|c|c|c|c|c|} 
\hline
Layer          & Size-in & Size-out & Layer details      & Parameters  \\ 
\hline
Concat         & 2×512   & 1024×1   &                    & 0           \\ 
\hline
Dense3         & 1024×1  & 512×1    & 512 neurons (relu) & 524800      \\ 
\hline
Batch norm4    & 512×1   & 512×1    &                    & 2048        \\ 
\hline
Dense4         & 512×1   & 256×1    & 256 neurons (relu) & 131,328     \\ 
\hline
Classification & 256×1   & 1        & Sigmoid            & 513         \\ 
\hline
Total          &         &          &                    & 18,172,577  \\
\hline
\end{tabular}
\end{table*}


\section{Experiments}
In this section, we conduct extensive experiments to evaluate the effectiveness of the proposed framework for representation learning for face spoofing detection in several datasets, which shows
the superiority of our proposed approach over the state-of-the-art baseline methods.

\subsection{Data Preparation and Preprocessing}

\paratitle{Face Detection.} Face detection is an essential step to properly deal with face anti-spoofing task. Due to its usefulness in many computer vision tasks, there exist several techniques to identify faces in images such as Haar Cascades, Viola-Jones and SSD~\cite{padilla2012evaluation,jones2003fast,zhang2017s3fd}. These methods have been used in previous research and worked sufficiently. However, in our work we decided to use Multi-task Cascade Convolutional Network (MTCNN) which has the ability of aligning faces in different poses and has adequate performance in unconstrained environments and various illuminations. MTCNN consists of three CNN networks leveraging a cascade architecture. The first stage detects the candidate facial windows and their bounding boxes and merges highly overlapping ones. In second and third stages the results are refined more and non-maximum suppression (NMS) is applied to filter out false candidates. Finally five facial landmark positions are obtained. We applied the method to the frames of videos to extract face images. Since in anti-spoofing task the background detail and information may be of great importance, a margin of pixels preserved around detected faces.

\paratitle{Color Texture Feature Extraction.} The value of color texture descriptors for face anti-spoofing detection have been proved by retrospective studies. In this part our method is mostly based on~\cite{boulkenafet2016face}. The main idea is that the artifact face image may suffer from different types of quality loss because of being affected by different camera systems and a display device such as mobile devices, monitors or papers. Therefore, the spoofed image can be detected by analyzing color texture features on channels of different color spaces such as HSV and YCbCr. The HSV and YCbCr color space has been proven useful in previous works due to the chrominance and luminance information which are less correlated than RGB channels. More details of the effectiveness of the color textures usages in PAD and color spaces differences can be found in~\cite{boulkenafet2016face,de2012lbp,maatta2011face}. 

To leverage this information, we have constructed our image representing vector from three feature descriptors: Local Binary Pattern (LBP), Co-occurrence of Adjacent Local Binary pattern (CoALBP) and Local Phase Quantization (LBQ) which are extracted from gray-scale image, HSV and YCbCr channels ( six descriptors in total), as described in the following. 

\emph{Local Binary Pattern (LBP)}: 
The Local Binary Pattern descriptor which is proposed in~\cite{ojala2000gray} is a gray-scale texture descriptor. Because of its discriminative power and computational simplicity, LBP has become a popular approach in various applications. To do so, a binary code is computed for each pixel by setting a Threshold for circularly symmetric neighborhood of adjacent pixels with the value of the central pixel, which can be stated as,    

\begin{align}
LBP_{P,R}(x,y) = \sum_{n=1}^{P} \delta( r_n \Vert -r_c) \times 2^{(n-1)}. 
\end{align}

where $\delta(x) = 1$ if $x>=0$, otherwise $\delta(x) = 0$. The intensity value of central pixel $(x,y)$ and its $P$ neighbor pixels in the circle of radius $R$, are denoted by $r_c$ and $r_n$ $(n = 1, \ldots, P)$, respectively.  Then the histogram is computed to measure the occurrence of different binary patterns.

\emph{Co-occurrence of Adjacent Local Binary Patterns (CoALBP):} 
In the LBP method, the information of spatial relation between patterns are not taken into account by the histogram. In order to take advantage of this information, the Co-occurrence of Adjacent Local Binary Patterns (CoALBP) is proposed in~\cite{nosaka2011feature}. After the LBP pattern are extracted from the image, four direction are defined such as $ D= \{ (0, \Delta d), (\Delta d, 0), (\Delta d, \Delta d), (-\Delta, \Delta d) \}$ exploit the correlation and similarity between the adjacent patterns, where $\delta d$ is the distance between two adjacent patterns. A $2$-dimensional histogram with size of $16 \times 16$ is created for each direction and the obtained histograms are concatenated to form the CoALBP descriptor~\cite{boulkenafet2016face}.

\emph{Local Phase Quantization (LPQ):}      
The Local Phase Quantization (LPQ) descriptor is mainly exploited to extract the texture information from the blurred images~\cite{ojansivu2008blur}. The Short Term Fourier Transform (STFT) is performed on image pixels to analyze the $M \times M$ neighborhoods adjacent to a center pixel $x$. Let $\mathcal{F}_u(x)$  be the STFT at the pixel $x$ using the bi-dimensional spatial frequency $u$. In the LPQ descriptor, only four complex frequencies are used: $u_0 = (\alpha, 0)$, $u_1 = (\alpha, \alpha)$, $u_2 = (0, \alpha)$, and $u_3 = (-\alpha, -\alpha)$, where $\alpha$ is a small scalar $(\alpha \ll 1)$, corresponding to the $0$, $45$ and $90$ and $135$ directions. To compute the LBQ descriptor, the LPQ features at a pixel $x$ are given by the vector, 
\begin{align}
    F_x = [ Re{F_{u0}(x), F_{u1}(x), F_{u2}(x), F_{u3}(x)}, \\ \nonumber Im{F_{u0}(x), F_{u1}(x), F_{u2}(x), F_{u3}(x)} ],
\end{align} 
where $Re{.}$ and $Im{.}$ are the real and the imaginary parts of a complex number, respectively. The elements of the vector $F_x$ are fed to the previously defined $\delta$ function to be binarized, then the resulting coefficients are represented as integer values in range of $0$ to $255$ and stored into a histogram. Also a de-correlation step based on the whitening transform before the quantization process is suggested to make the LPQ coefficients statistically independents~\cite{boulkenafet2016face}. Table~\ref{tbl::shallow-features} illustrates the details of the different descriptors used in this work.


\subsection{Experimental Setting}
The proposed framework is designed to examine selfie videos submitted by users to a biometric system in order to detect spoofed faces for illegitimate access. To do so, the videos are passed into several steps. First, to organize the data, all video frames are extracted and MTCNN face detection is applied to them. A margin of $44$ pixels is also added to help detecting artifacts cues that may be existed in the background image. Then, the cropped image is resized to $160 \times 160$. To accelerate the training process, first we produced the data for both channels before feeding to the network. Thus, the normalized RGB image is converted to HSV, YCbCr and gray-scale color spaces and the texture descriptors are extracted from each channels of HSV, YCbCr and gray-scale separately and concatenated to form the enhanced feature vector. The resulting vector is used as an input for Wide channel. The parameters of each feature are provided in Table~\ref{tbl::shallow-features}. Finally, the face images and their corresponding feature vectors are fed into the network which was explained in section~\ref{sec::framework}. Also, it is worth noting that the learning rate and decay coefficient are 0.001 and the momentum term is $0.9$. The output value of the network describes the probability of spoofing attack in the image.

\begin{figure*}
  \includegraphics[width=\columnwidth]{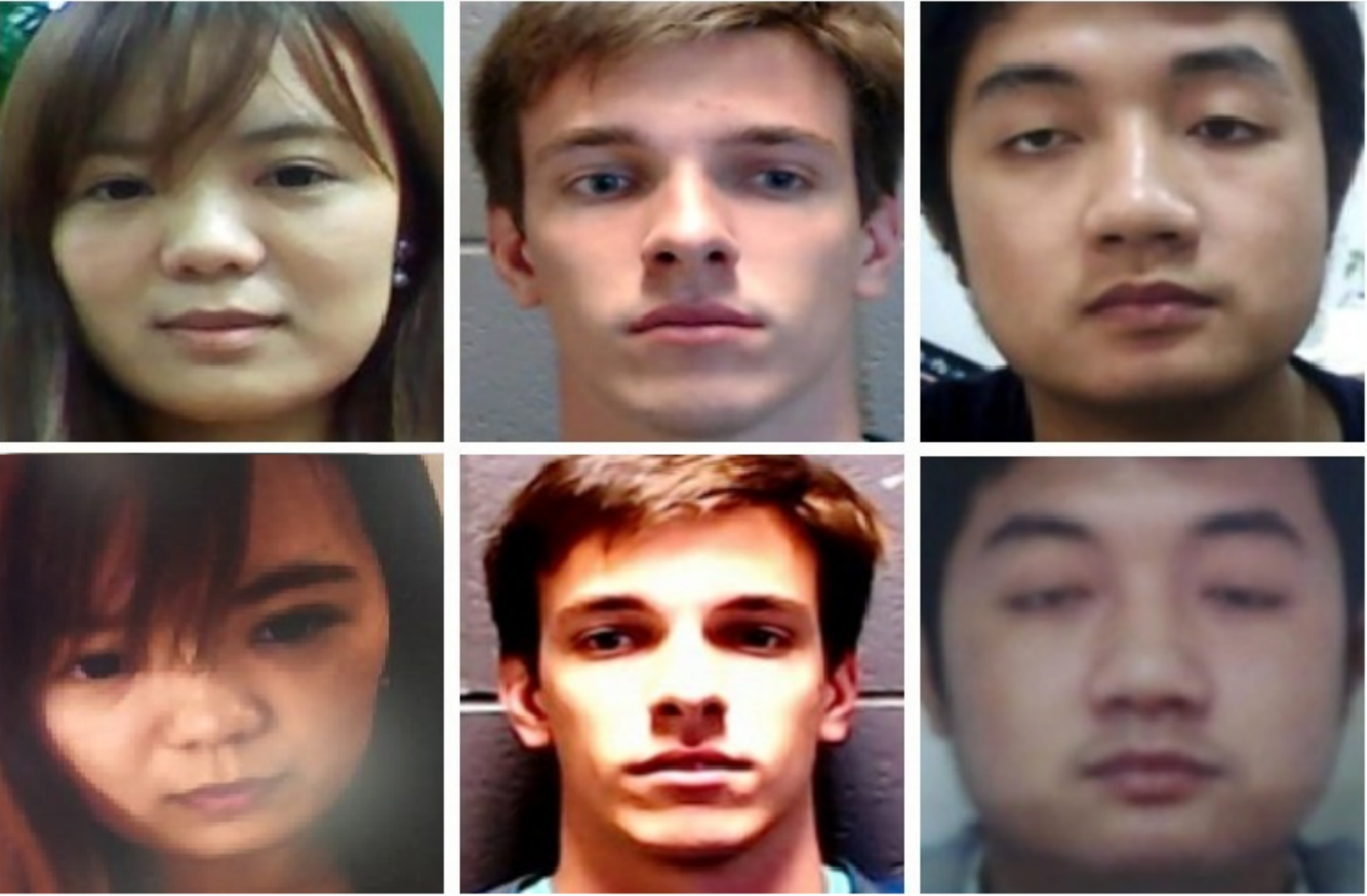}
  \caption{Samples of real and fake images in Rose-Youtu, SiW and NUAA datasets}
  \label{fig::framework}
\end{figure*}




\subsection{Datasets}
We verified the performance of our proposed method on three widely-used datasets: SiW, ROSE-Youtu and NUAA Imposter. In the first two dataset different spoofing attacks using different devices are recorded. NUAA, however, only focuses on print attacks. All the datasets have already been divided to training and development sets and we used the same settings for the experiments. A short description of each dataset is brought in the following.

\paratitle{SiW dataset}:  SiW includes live and spoof videos from $165$ subjects, where, for each subject, there are $8$ live and up to $20$ spoof videos, in total $4,478$ videos~\cite{liu2018learning}. All videos are captured in $30$ fps, about $15$ second length. The recording resolution is $1080P$ HD. The live videos are collected in four sessions with variations of distance, pose, illumination and expression. The spoof videos are collected with various attacks such as printed paper and replay.

\paratitle{ROSE-Youtu dataset}: This dataset covers a large variety of illumination conditions, camera models, and attack types, consists of $3350$ videos with $20$ subjects publicly available~\cite{li2018unsupervised}. For each subject, there are $150$ to $200$ video clips with the average duration around 10 seconds recorded by 5 different mobile phones with different resolutions. There exist $25$ genuine face video which covers $5$ different illumination conditions in office environment, including clients wearing eye-glasses. They considered three spoofing attack types including printed paper attack, video replay attack, and mask attack.

\paratitle{NUAA Imposter dataset}: The dataset consists of 15 subjects videos captured by webcams~\cite{tan2010face}. The photos of both genuine subjects and their spoofing images are recorded with a frame rate of 20 fps. Five hundred images are gathered for each. Pictures are all frontal with a neutral expression. Location and illumination condition of each session are varied.


\subsection{Results}
In order to evaluate the performance of our approach, we trained our model on SiW and ROSE-Youtu training parts separately. The evaluation is done on the test parts of these datasets with EER (equal error rate) and HTER (half total error rate) metrics. Also to measure the generalization, the cross dataset test is done for both of them (training on SiW and testing with ROSE-Youtu and vice versa). Since NUAA is small dataset and limited to print attack, it is used just for evaluations on both testing scenarios. The results are illustrated in table 5.

\begin{table}[]
\centering
\caption{Proposed model accuracy on cross-dataset settings}
\label{tbl::performance}
\begin{tabular}{|l|c|c|c|c|c|c|}
\hline
\textbf{\begin{tabular}[c]{@{}l@{}}Train/Test\\ Datasets\end{tabular}} & \multicolumn{2}{c|}{\textbf{SiW}} & \multicolumn{2}{c|}{\textbf{ROSE}} & \multicolumn{2}{c|}{\textbf{NUAA}} \\ \hline
\textbf{Metric}                                                        & \textbf{EER}    & \textbf{HTER}   & \textbf{EER}    & \textbf{HTER}    & \textbf{EER}    & \textbf{HTER}    \\ \hline
\textbf{SiW}                                                           & 0.55\%          & 1.05\%          & 19.57\%         & 25.21\%          & 26.57\%         & 29.22\%          \\ \hline
\textbf{ROSE}                                                          & 17.83\%         & 23.37\%         & 4.27\%          & 6.12\%           & 22.34\%         & 27.48\%          \\ \hline
\end{tabular}
\end{table}

\begin{table*}[h]
\centering
\caption{Separate channels evaluation on cross-dataset test}
\label{tbl::separate-channel-evaluation}
\begin{tabular}{|l|c|c|c|c|c|c|c|c|}
\hline
\textbf{\begin{tabular}[c]{@{}l@{}}Train/Test\\ Datasets\end{tabular}} & \multicolumn{4}{c|}{\textbf{SiW}}                                                                                         & \multicolumn{4}{c|}{\textbf{ROSE}}                                                                                      \\ \hline
\textbf{Method}                                                        & \multicolumn{2}{c|}{\textbf{CNN}}                           & \multicolumn{2}{c|}{\textbf{Color-Texture}}                 & \multicolumn{2}{c|}{CNN}                                   & \multicolumn{2}{c|}{\textbf{Color-Texture}}                \\ \hline
\textbf{Metric}                                                        & \textbf{EER}                 & \textbf{HTER}                & \textbf{EER}                 & \textbf{HTER}                & \textbf{EER}                & \textbf{HTER}                & \textbf{EER}                & \textbf{HTER}                \\ \hline
\textbf{SiW}                                                           & 1.05\%                       & 1.68\%                       & 1.92\%                       & 3.41\%                       & 28.57\%                     & 34.33\%                      & 34.75\%                     & 39.67\%                      \\ \hline
\textbf{ROSE}                                                          & \multicolumn{1}{l|}{27.51\%} & \multicolumn{1}{l|}{32.02\%} & \multicolumn{1}{l|}{30.70\%} & \multicolumn{1}{l|}{35.82\%} & \multicolumn{1}{l|}{9.34\%} & \multicolumn{1}{l|}{10.23\%} & \multicolumn{1}{l|}{8.95\%} & \multicolumn{1}{l|}{10.81\%} \\ \hline
\end{tabular}
\end{table*}

It can be inferred from the results that while ROSE-Youtu is a smaller than SiW, it is more reliable and general dataset and produced more generalized results in cross dataset test. On the other hand SiW is large and more biased dataset. Because of lower quality images of NUAA which increases FRR (False rejection rate), results drops greatly in EER.
Furthermore, it is important to measure how much improvement is achieved by using this approach compare to each single channel networks. to compare the performance over each channel, the feature interaction block is replaced with sigmoid classification layer and the network is trained with the same data as before. The comparing results is shown in following Table6.

It can be seen in the table that the dual channel approach outperforms both single channel models. It is also produced more generalized model with better results in cross dataset tests. Also, while the descriptor model can perform pretty well on one dataset, it is more prone the get overfit when facing an unseen dataset and has lower results on cross dataset test, while CNN model can still perform better extracting general features.

\section{Conclusion and Future Work}

In this paper we proposed a dual channel method to use both CNN and color texture descriptors domains. According to the results our method can not only make a significant improvement comparing to previous similar methods, but also It comes out it can be a effective method to extract well-generalized and robust features to use for cross dataset experiments and avoid biases between datasets. For the future, one useful lead is using transfer learning for the CNN channel with more sophisticated architectures. Also other descriptors can be added to the other channel to derive better representation for anti-spoofing problem.

